%% file: main.tex
\documentclass[conference]{IEEEtran}
\IEEEoverridecommandlockouts
\usepackage{graphicx}
\usepackage[utf8]{inputenc}
\usepackage{booktabs}
\usepackage[utf8]{inputenc}
\usepackage{geometry}
\usepackage{graphicx}
\usepackage{multirow}
\usepackage{amsmath}
\usepackage{amsthm}
\usepackage{stackengine}
\usepackage{amssymb}
\usepackage{algorithm}
\usepackage{algorithmicx}
\usepackage{algpseudocode} 
\usepackage{makecell}
\usepackage{color}
\usepackage{tikz}
\usepackage{caption}
\usepackage{lscape}
\usepackage{pdflscape}
\usepackage{rotating}
\usepackage{algpseudocode}
\usepackage{empheq} 
\usepackage{color,soul}
\usepackage{hyperref}

\usepackage{subcaption}

\usepackage{amsmath}
\usetikzlibrary{shapes.arrows,arrows.meta}
\usetikzlibrary{positioning, automata, chains, arrows.meta}
\usetikzlibrary{matrix,positioning,calc}

\algnewcommand{\LineComment}[1]{\State \(\triangleright\) #1}

\hypersetup{
    colorlinks=true,
    linkcolor=blue,
    filecolor=blue,      
    urlcolor=blue,
    citecolor=blue
}

\def\BibTeX{{\rm B\kern-.05em{\sc i\kern-.025em b}\kern-.08em
    T\kern-.1667em\lower.7ex\hbox{E}\kern-.125emX}}
\begin{document}

\title{Learning Minimalistic Tsetlin Machine Clauses with Markov Boundary-Guided Pruning\thanks{979-8-3503-4477-6/23/\$31.00 ©2023 IEEE}}

\author{\IEEEauthorblockN{Ole-Christoffer Granmo}
\IEEEauthorblockA{\textit{Centre for AI Research} \\
\textit{University of Agder}\\
Grimstad, Norway \\
0000-0002-7287-030X}
\and
\IEEEauthorblockN{Per-Arne Andersen}
\IEEEauthorblockA{\textit{Centre for AI Research} \\
\textit{University of Agder}\\
Grimstad, Norway \\
0000-0002-5490-6436}
\and
\IEEEauthorblockN{Lei Jiao}
\IEEEauthorblockA{\textit{Centre for AI Research} \\
\textit{University of Agder}\\
Grimstad, Norway \\
0000-0002-5490-6436}
\and
\IEEEauthorblockN{Xuan Zhang}
\IEEEauthorblockA{\textit{Energy and Technology} \\
\textit{NORCE}\\
Grimstad, Norway\\
xuzh@norceresearch.no}
\and
\IEEEauthorblockN{Christian Blakely}
\IEEEauthorblockA{\textit{Machine Learning and AI} \\
\textit{PwC Switzerland}\\
Zurich, Switzerland\\
christian.blakely@pwc.ch}
\and
\IEEEauthorblockN{Geir Thore Berge}
\IEEEauthorblockA{\textit{Technology and Innovation} \\
\textit{Sørlandet Hospital}\\
Kristiansand, Norway\\
0000-0002-7631-8045}
\and
\IEEEauthorblockN{Tor Tveit}
\IEEEauthorblockA{\textit{Technology and Innovation} \\
\textit{Sørlandet Hospital}\\
Kristiansand, Norway\\
tor.tveit@sshf.no}
}

\maketitle

\begin{abstract}
A set of variables is the Markov blanket of a random variable if it contains all the information needed for predicting the variable. If the blanket cannot be reduced without losing useful information, it is called a Markov boundary. Identifying the Markov boundary of a random variable is advantageous because all variables outside the boundary are superfluous. Hence, the Markov boundary provides an optimal feature set. However, learning the Markov boundary from data is challenging for two reasons. If one or more variables are removed from the Markov boundary, variables outside the boundary may start providing information. Conversely, variables within the boundary may stop providing information. The true role of each candidate variable is only manifesting when the Markov boundary has been identified. In this paper, we propose a new Tsetlin Machine (TM) feedback scheme that supplements Type I and Type II feedback. The scheme introduces a novel Finite State Automaton --- a Context-Specific Independence Automaton. The automaton learns which features are outside the Markov boundary of the target, allowing them to be pruned from the TM during learning. We investigate the new scheme empirically, showing how it is capable of exploiting context-specific independence to find Markov boundaries. Further, we provide a theoretical analysis of convergence. Our approach thus connects the field of Bayesian networks (BN) with TMs, potentially opening up for synergies when it comes to inference and learning, including TM-produced Bayesian knowledge bases and TM-based Bayesian inference.
\end{abstract}


\section{Introduction}

The power of state-of-the-art machine learning comes from encoding enormous amounts of historical data using millions (and lately, trillions) of parameters. However, the historical data makes the trained models carry on biases, discrimination, and prejudices~\cite{Bender2021Parrots}, while the number of parameters makes them incomprehensible to humans~\cite{Rudin2019}.

Researchers are further accumulating evidence that models based on correlation are brittle. Even state-of-the-art deep learning models, with their high computational cost and carbon footprint~\cite{Schwartz2020GreenA}, tend to learn simple correlations instead of capturing the underlying causal dynamics of the data~\cite{scholkopf2021causal,sauer2021counterfactual}. Relying on such correlations is problematic when they are spurious or accurate only in limited contexts due to data bias.

Recently, the emerging paradigm of TMs~\cite{granmo2018tsetlin} has made a fundamental shift from arithmetic-based to logic-based machine learning. Seen from a logical engineering~\cite{lucas1995} perspective, a TM produces propositional/relational clauses in Horn form (logical AND-rules)~\cite{Saha2022}. However, the logical expressions are robustly learnt using finite state machines, so-called Tsetlin automata \cite{Tsetlin1961}. TMs handle uncertainty despite being based on logic, using multiple clauses to signify confidence~\cite{abeyrathna2020confidence}.  In this way, TMs introduce the concept of logically interpretable learning, where both the learned model and the process of learning are easy to follow and explain.

Two different feedback types interact to make the TM learn patterns: Type~I and Type~II Feedback. Type~I Feedback produces frequent patterns, leading to descriptive conjunctive clauses. To increase discrimination power, the TM simultaneously uses Type~II Feedback to introduce discriminative literals into the clauses. However, while effective in pattern recognition, the clauses sometimes include more context than needed, reducing efficiency and potentially interpretability.

To increase the efficiency and interpretability of TMs, we here propose a new kind of TM feedback -- \emph{Type~III Feedback}. Type~III Feedback interacts with Type I and Type II Feedback, with the goal of isolating the mechanism that governs the target variable. That is, Type~III Feedback prunes the clauses of literals so that only the literals that directly govern the target variable remain.

To reach the above goal, our strategy is to view the TM clauses from a BN ~\cite{pearl1988} perspective, introducing the concepts of \emph{Markov blankets} and \emph{Markov boundaries}~\cite{JMLR:v19:14-033} into TM learning.

\textbf{Markov Blanket.} In BNs, the Markov blanket of a target variable is a subset of the other variables. When we know the value of the variables in this subset, the remaining variables become superfluous. That is, the variables outside the Markov blanket are conditionally independent of the target given the variables in the Markov blanket.

\textbf{Markov Boundary.} The Markov boundary is the smallest possible Markov blanket. In other words, none of its subsets are Markov blankets. As proven by Pellet and Elisseeff, the Markov boundary of a target is the theoretically optimal set of features for predicting that target~\cite{JMLR:v9:pellet08a}.

\textbf{Context-Specific Independence.} Context-specific independence is independence that holds in certain contexts~\cite{boutilier1996}. In brief, the independence holds only for specific variable values. Such a refinement allows us to distil the mechanism governing the target value in more detail, leading  to context-specific Markov blankets.

\textbf{Context-Specific Markov Blankets.} Context-specific Markov blankets exploit context-specific independence. The purpose is to reduce the number of parameters needed to specify the factors deciding the target variable~\cite{klein2004}.

\textbf{Paper Contributions.} This paper introduces a novel feedback scheme that exploits context-specific independence to prune clauses of literals outside their context-specific Markov boundary. The feedback scheme is based on a novel Context-Specific Independence Automaton (CS-IA) that works alongside the Tsetlin automata of the TM. As soon as an IA uncovers context-specific independence in a clause, it starts pruning any independent literals from the clause. We show empirically and prove formally that the CS-IA is capable of uncovering context-specific independence through on-line learning. We also show that CS-IA can robustly reduce the number of literals used on MNIST.

\begin{figure}[ht]
\centering
\includegraphics[width=0.7\columnwidth]{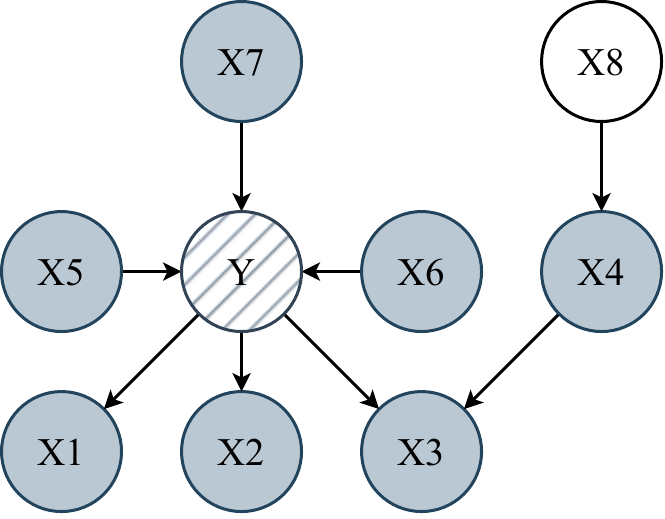}
\caption{Markov boundary of $Y$ in grey.}\label{fig:BN}
\end{figure}

\section{Context-Specific Markov Boundary-Guided Pruning}

\subsection{Tsetlin Machine Clauses and Bayesian Networks}

\subsubsection{Bayesian Networks} 
A BN is a directed acyclic graph (DAG) where the nodes of the graph are random variables while the directed edges represent dependencies among the variables. The entire graph is a compact representation of a joint probability distribution over the set of random variables, where each node and its parents are associated with a conditional probability distribution (CPD). In this manner, the CPD gives a probabilistic formulation of the relationship between two or more variables. For example, Table \ref{tab:cpd-x1} contains the CPD for variable $X_1$ of the BN in Fig. \ref{fig:BN}.  In this example, the parent node of $X_1$ is $Y$. The table therefore specify a probability distribution over $X_1$, conditioned on the values of $Y$. When $Y=0$, $X_1$ takes the value $0$ with probability $0.9$. Conversely, it takes value $1$ with probability $0.1$. Note that the variables of a BN can also be continuous, with relationships modelled by conditional density functions. 

Learning a BN from data is conducted in two phases. The first phase creates the structure of the network and the second phase estimates the probabilities of the CPDs, given the network structure from phase one. While parameter estimation is a well-studied problem, structure learning is still considered challenging. Vanilla BN structure learning usually involves two strategies: 1) constraint-based search for conditional independencies (CIs) in the data, from which one builds a DAG that is consistent with these; and 2) score-based search which poses structure learning as an optimization problem. One then seeks to maximize the score over the space of possible DAGs. Finding an optimal structure, however, has been shown to be NP-hard. 

\subsubsection{Tsetlin Machine Clauses}

A TM learns conjunctive clauses to represent patterns in the data. A TM clause is simply an AND-rule for predicting a target value. For instance, a TM could represent the relationship between $X_1$ and the target $Y$ with the two clauses:
\begin{eqnarray}
C^+(\mathbf{X})&=& X_1\label{eqn:example_clause},\\
C^-(\mathbf{X})&=& \lnot X_1.
\end{eqnarray}
Each clause takes a vector $\mathbf{X} = [X_1, X_2, \ldots, X_8]$ as input. Half of the TM clauses get positive polarity, predicting when the target is $1$. The clause $C^+$ above has positive polarity (signified by the upper index), and predicts $Y = 1$ when $X_1 = 1$. Relating the clause to the CPD, it has precision $0.9$ when it comes to predicting when $Y=1$. The negative polarity clauses, on the other hand, predicts when $Y=0$. For instance, the clause $C^-$ above predicts $Y=0$ when $X_1=0$, again giving precision $0.9$ according to the CPD.

A detailed description of how the TM uses teams of clauses to learn patterns from data can be found in \cite{granmo2018tsetlin}. The focus here is on the new learning automaton that we introduce to learn context-specific Markov boundaries. 

\subsubsection{Context-Specific Independence and Clauses} By way of example, two variables $X_5$ and $X_7$ are \emph{independent} if we have $P(X_5, X_7) = P(X_5)P(X_7)$. Further, the variables $X_1$ and $X_2$ are \emph{conditionally independent} given $Y$ if we have $P(X_1, X_2 | Y) = P(X_1 | Y) P(X_2 | Y)$. \emph{Context-specific independence} refines conditional independence by saying that two variables $X_1$ and $X_2$ are independent when another variable $Y$ takes a specific value, for instance $0$, but not necessarily for other $Y$-values. That would be the case if we have $P(X_1, X_2 | Y = 0) = P(X_1 | Y = 0) P(X_2 | Y = 0)$. In the latter sense, one can say that a TM clause defines a specific context. For instance, the clause $C^+$ in Eqn.~\ref{eqn:example_clause} specifies a context where $X_1=0$.

\subsubsection{Context-Specific Markov Boundaries and Clauses} A \emph{Markov blanket} of a target variable $Y$ is a subset $\mathcal{X}' \subseteq \{X_1, X_2, \ldots, X_8\}$ of the remaining variables $\mathcal{X} = \{X_1, X_2, \ldots, X_8\}$ where all the variables outside $\mathcal{X}$ are conditionally independent of the target $Y$: $P(Y | \mathcal{X}', \mathcal{X} \setminus \mathcal{X}') = P(Y | \mathcal{X}')$. The \emph{Markov boundary} of the target $Y$ is a Markov blanket that does not contain any smaller Markov blankets. As such, the Markov boundary of the target contains all the information needed for predicting the target. It is the theoretically optimal feature set~\cite{JMLR:v9:pellet08a}.  So-called \emph{d-separation} from \cite{pearl1988} gives that the Markov boundary of $Y$ in Fig. \ref{fig:markov-boundary-toy} is $X_1, X_2, \ldots, X_7$, rendering $X_8$ conditionally independent of $Y$.  When the Markov  boundary only applies for specific values of $X_1, X_2, \ldots, X_7$, we say that the Markov boundary is context-specific. Again, a TM clause specifies a particular context by assigning values to the variables in the clause. For instance, the clause $C^+ = X_1 \land \lnot X_3$ specifies the context where we have both $X_1=1$ and $X_3=0$. 

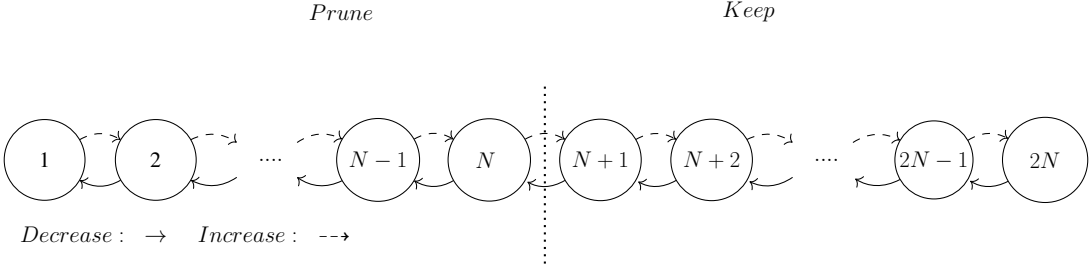
\begin{figure*}[!!h]
\centering
\resizebox{1\textwidth}{!}{
\begin{minipage}{1\textwidth}
\begin{tikzpicture}[node distance = .35cm, font=\Huge]
    \tikzstyle{every node}=[scale=0.35]
    \node[state] (A) at (0,2) {~~~~1~~~~~};
    \node[state] (B) at (1.5,2) {~~~~2~~~~~};
    
    \node[state,draw=white] (M) at (3,2) {~~~$....$~~~};
    
    \node[state] (C) at (4.5,2) {$~N-1~$};
    \node[state] (D) at (6,2) {$~~~N~~~~$};
    
    \node[state] (E) at (7.5,2) {$~N+1~$};
    \node[state] (F) at (9,2) {$~N+2~$};
    
    \node[state,draw=white] (G) at (10.5,2) {~~~$....$~~~};
    
    \node[state] (H) at (12,2) {$2N-1$};
    \node[state] (I) at (13.5,2) {$~~~2N~~~$};

    \node[thick] at (4,4) {$Prune$};
    \node[thick] at (9.5,4) {$Keep$};
    
    \node[thick] at (1.9,1) {$Decrease:~\rightarrow$~~~~$Increase:~\dashrightarrow$};

    \draw[dashed,->]
    (A) edge[bend left] node [scale=1.2, above=0.1 of B]{} (B)
    (B) edge[bend left] node  [scale=1.2, above=0.1 of M] {} (M)
    (M) edge[bend left] node  [scale=1.2, above=0.1 of C] {} (C)
    (C) edge[bend left] node [scale=1.2, above=0.1 of D] {} (D)
    (D) edge[bend left] node  [scale=1.2, above=0.1 of E] {} (E);

    \draw[every loop]
    (I) edge[bend left] node [scale=1.2, below=0.1 of H] {} (H)
    (H) edge[bend left] node  [scale=1.2, below=0.1 of G] {} (G)
    (G) edge[bend left] node [scale=1.2, below=0.1 of F] {} (F)
    (F) edge[bend left] node  [scale=1.2, below=0.1 of E] {} (E)
    (E) edge[bend left] node  [scale=1.2, below=0.1 of D] {} (D);

    
    \draw[->]
    (B) edge[bend left] node  [scale=1.2, above=0.1 of A] {} (A)
    (M) edge[bend left] node [scale=1.2, above=0.1 of B] {} (B)
    (C) edge[bend left] node [scale=1.2, above=0.1 of M] {} (M)
    (D) edge[bend left] node [scale=1.2, above=0.1 of C] {} (C);
    
    \draw[dashed,->]

    (H) edge[bend left ] node [scale=1.2, below=0.1 of I] {} (I)
    (G) edge[bend left] node  [scale=1.2, below=0.1 of H] {} (H)
    (F) edge[bend left] node  [scale=1.2, below=0.1 of G] {} (G)
    (E) edge[bend left ] node [scale=1.2, below=0.1 of F] {} (F);
    
      \draw[dotted, thick] (6.75,0.6) -- (6.75,3);

\end{tikzpicture}
\end{minipage}
}
\caption{The Context-Specific Independence Automaton (CS-IA) with $2N$ states.}
\label{figure:TAarchitecture_basic}
\end{figure*}

\begin{figure*}

\resizebox{1\textwidth}{!}{
\begin{minipage}{1\textwidth}
\begin{tikzpicture}[font=\sffamily,bullet/.style={fill=black,circle,text=white,font=\sffamily\bfseries,inner sep=1.2pt}]
\draw[thick,-{Triangle[open]}] (-0.2,0) -- (14,0);
\foreach \X in {1,2,3,4}
{\draw ({(\X-1)*4},0.1) -- ({(\X-1)*4},-0.1);
\draw[thick,dashed] ({(\X-1)*4},0) -- ++(0,1)
node[above,bullet]{\X};}
\foreach \y in {5,6,7,8}
{\draw ({(\y-1)*1.5},0.1) -- ({(\y-1)*1.5},-0.1);
\draw[thick,red] ({(\y-1)*1.5},0) -- ++(0,1)
node[above,bullet]{\y};}
\node[anchor=north] at (0,0) {0};
\node[anchor=north east] at (14,0) {Time};
\end{tikzpicture}
\end{minipage}
}
\caption{Illustration of the meaning of ``Wait" in the learning loop.}
\label{figure:illustration}
\end{figure*}
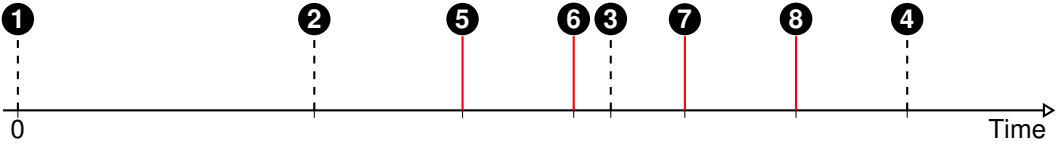

\subsection{Context-Specific Independence Automaton (CS-IA)} \label{newLA}

Looking at TM clauses from the viewpoint of context-specific Markov blankets, we hypothesize that Type~I and Type~II Feedback sometimes may include superfluous variable assignments. E.g., they may potentially include $X_8$ in Fig. \ref{fig:BN} in clauses predicting $Y=1$ even though $X_8$ is conditionally independent of $Y$ given $X_1, X_2, \ldots, X_7$. The conditionally independence means that $P(Y| X_1, X_2, \ldots, X_7, X_8) = P(Y| X_1, X_2, \ldots, X_7)$, rendering $X_8$ superfluous for predicting $Y$. By allowing the TM to discover that $X_8$ is superfluous, it can create shorter clauses that more concisely capture the underlying dynamics in the data.

To this end, we introduce the Context-Specific Independence Automaton (CS-IA), illustrated in Fig. \ref{figure:TAarchitecture_basic} and detailed below. The CS-IA is based on the Tsetlin automaton~\cite{Tsetlin1961}, being a finite state automaton with two actions. In brief, like with standard Type I and Type II Feedback, each clause literal gets its own automaton.

\subsubsection{CS-IA Actions} Action \emph{Prune} means that the CS-IA decides to remove its literal from its clause because it has discovered that the literal is conditionally independent of the target given the remaining literals in the clause (the context).

\subsubsection{CS-IA State Initialization and Updating} 

By way of example, we exploit the fact that  $P(Y| X_1, X_2, \ldots, X_7, X_8) = P(Y| X_1, X_2, \ldots, X_7)$ when $X_8$ is outside the Markov boundary of $Y$. In brief, our CS-IA is to learn to Prune the literal if it is confident that $P(Y| X_1, X_2, \ldots, X_7, X_8) = P(Y| X_1, X_2, \ldots, X_7)$. To avoid premature pruning of literals, the CS-IA starts in state $2N$ (extreme right in the figure), signifying high confidence in dependence.

After initialization, the CS-IA swaps between examining two scenarios to learn context-specific independence for its clause $C$ from data:
\begin{enumerate}
    \item The clause $C$ evaluates to $1$ and $X_8$ takes part in $C$.
    \item The clause $C$ evaluates to $1$ and $X_8$ takes part in $C$, however, we pretend that $X_8$ is not there.
\end{enumerate}
This means that the CS-IA waits for Scenario 1 to happen, then for Scenario 2, and then for Scenario 1 again, and so on. Fig. \ref{figure:illustration} illustrates this waiting strategy, where the dotted line refers to Scenario $1$ and the solid red line to Scenario 2. Only Event 2, Event 5, Event 3, and Event 8 are used for updating. Using the above procedure, it will be possible for the CS-IA to gradually uncover whether $P(Y| X_1, X_2, \ldots, X_7, X_8) = P(Y| X_1, X_2, \ldots, X_7)$, because if they are equal, Scenario 1 and Scenario 2 will look the same when it comes to predicting $Y$.

The pseudo-code for updating the state and taking actions for a general variable $X_i$ with Independence Automaton $\mathit{IA}_i$ is found below (see the next two sections for an empirical study and a rigorous theoretical convergence analysis of the scheme): 
\begin{enumerate}
\item Wait for $X_i$=1 and $C$=1:
\begin{enumerate}
\item If $Y = 1$ then increase the state of $\mathit{IA}_i$.
\item If $Y = 0$ then decrease the state of $\mathit{IA}_i$.
\item Decrease the state of $\mathit{IA}_i$ with a small probability $1.0/d$ ($d$ is a hyperparameter that controls how conservative the pruning should be).
\end{enumerate}
\item Wait for the remaining variables of $C$ being 1, ignoring the value of $X_i$ (simulating that $X_i$ is not present in the clause).
\begin{enumerate}
\item If $Y = 1$ then decrease the state of $\mathit{IA}_i$.
\item If $Y = 0$ then increase the state of $\mathit{IA}_i$.
\item Decrease the state of $\mathit{IA}_i$ with probability $1.0/d$.
\item If state is below or equal to $N$, prune the variable $X_i$ from the clause $C$.
\end{enumerate}
\item Repeat.
\end{enumerate}

We coin the new feedback type Type III Feedback, working alongside Type I and Type II Feedback.

\section{Empirical Study}

\begin{figure*}
    \footnotesize
    \centering
    \begin{minipage}{0.25\textwidth}
        \centering
        \begin{tabular}{cc}
        \toprule
        X5(0)  & 0.6 \\
        X5(1)  & 0.4 \\
        \bottomrule
        \end{tabular}
        \captionof{table}{CPD of X5}
        \label{tab:cpd-x5}
    \end{minipage}\hfill
    \begin{minipage}{0.25\textwidth}
        \centering
        \begin{tabular}{cc}
        \toprule
        X6(0)  & 0.7 \\
        X6(1)  & 0.3 \\
        \bottomrule
        \end{tabular}
        \captionof{table}{CPD of X6}
        \label{tab:cpd-x6}
    \end{minipage}\hfill
    \begin{minipage}{0.25\textwidth}
        \centering
        \begin{tabular}{cc}
        \toprule
        X7(0)  & 0.8 \\
        X7(1)  & 0.2 \\
        \bottomrule
        \end{tabular}
        \captionof{table}{CPD of X7}
        \label{tab:cpd-x7}
    \end{minipage}\hfill
    \begin{minipage}{0.25\textwidth}
        \centering
        \begin{tabular}{cc}
        \toprule
        X8(0)  & 0.9 \\
        X8(1)  & 0.1 \\
        \bottomrule
        \end{tabular}
        \captionof{table}{CPD of X8}
        \label{tab:cpd-x8}
    \end{minipage}\hfill\vspace{0.5cm}
    \begin{minipage}{0.33\textwidth}
        \centering
        \begin{tabular}{ccc}
        \toprule
        Y      & Y(0)  & Y(1)  \\
        X1(0)  & 0.9   & 0.1   \\
        X1(1)  & 0.1   & 0.9   \\
        \bottomrule
        \end{tabular}
        \captionof{table}{CPD of X1}
        \label{tab:cpd-x1}
    \end{minipage}\hfill
    \begin{minipage}{0.33\textwidth}
        \centering
        \begin{tabular}{ccc}
        \toprule
        Y      & Y(0)  & Y(1)  \\
        X2(0)  & 0.8   & 0.6   \\
        X2(1)  & 0.2   & 0.4   \\
        \bottomrule
        \end{tabular}
        \captionof{table}{CPD of X2}
        \label{tab:cpd-x2}
    \end{minipage}\hfill
    \begin{minipage}{0.33\textwidth}
        \centering
        \begin{tabular}{ccc}
        \toprule
        X8     & X8(0)  & X8(1)  \\
        X4(0)  & 0.6    & 0.4    \\
        X4(1)  & 0.4    & 0.6    \\
        \bottomrule
        \end{tabular}
        \captionof{table}{CPD of X4}
        \label{tab:cpd-x4}
    \end{minipage}\hfill
    \vspace{0.5cm} 

    \begin{minipage}{1.0\textwidth}
      \centering
        \begin{tabular}{ccccccccc}
        \toprule
        X5    & X5(0)  & X5(0)  & X5(0)  & X5(0)  & X5(1)  & X5(1)  & X5(1)  & X5(1)  \\
        X6    & X6(0)  & X6(0)  & X6(1)  & X6(1)  & X6(0)  & X6(0)  & X6(1)  & X6(1)  \\
        X7    & X7(0)  & X7(1)  & X7(0)  & X7(1)  & X7(0)  & X7(1)  & X7(0)  & X7(1)  \\
        Y(0)  & 0.3    & 0.4    & 0.7    & 0.6    & 0.1    & 0.8    & 0.3    & 0.4    \\
        Y(1)  & 0.7    & 0.6    & 0.3    & 0.4    & 0.9    & 0.2    & 0.7    & 0.6    \\
        \bottomrule
        \end{tabular}
        \captionof{table}{CPD of Y}
         \label{tab:cpd-y}
    \end{minipage}
\end{figure*}

\begin{figure}
    \centering
    \includegraphics[width=1.0\linewidth]{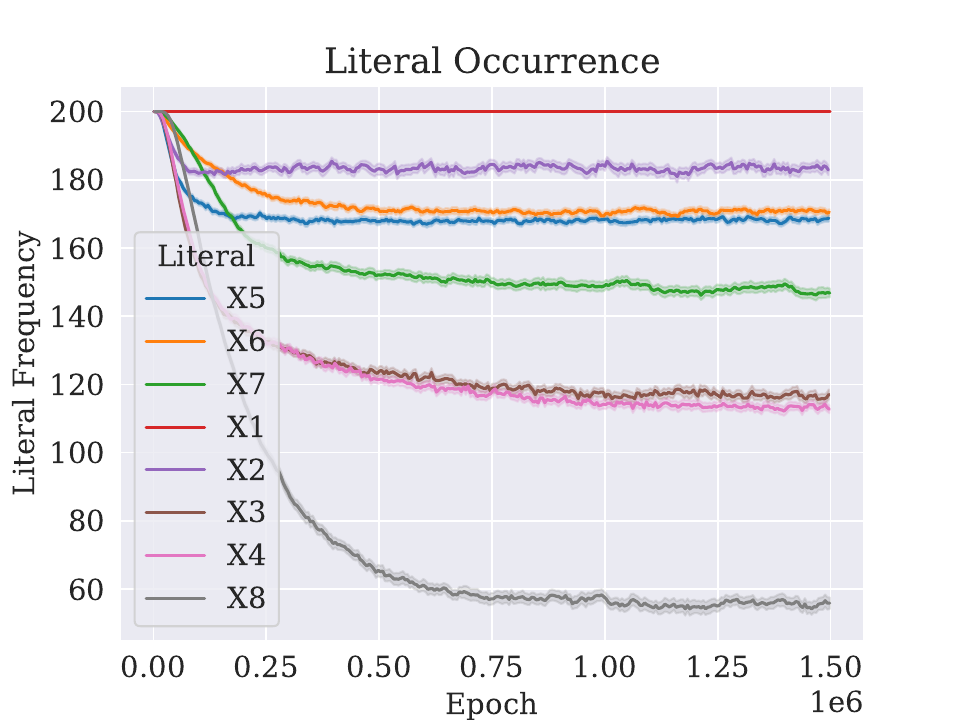}
    \caption{Each line represents the frequency at which a literal is present in a clause. Literals which occur less frequently are either not part of the Markov boundary or have substantial uncertainty bound to their outcome.}
    \label{fig:markov-boundary-literal-occurence}
\end{figure}

\begin{figure*}
    \centering
    \begin{subfigure}[b]{0.48\linewidth}
        \includegraphics[width=\linewidth]{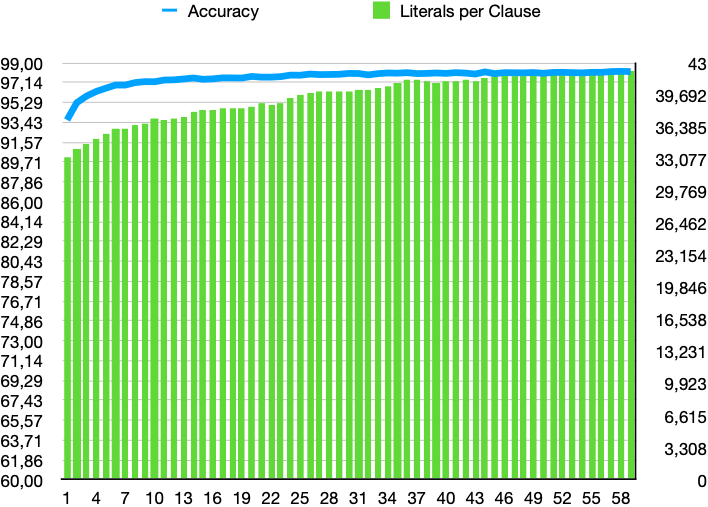}
        \caption{}
        \label{fig:mnist_without_pruning}
    \end{subfigure}
    \hfill
    \begin{subfigure}[b]{0.5\linewidth}
        \includegraphics[width=\linewidth]{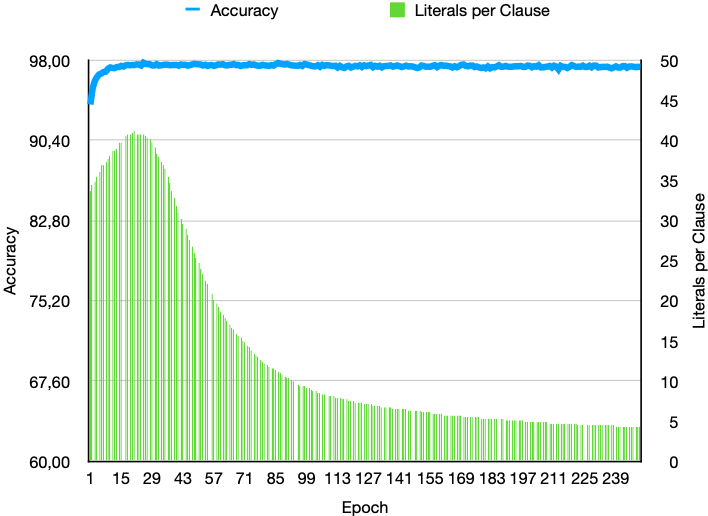}
        \caption{}
        \label{fig:mnist_with_pruning}
    \end{subfigure}
     \newline
    \caption{Accuracy (blue) and literals (green) without (a) and with (b) pruning.}
    \label{fig:mnist}
\end{figure*}

\begin{table*}
    \centering
    \input{Figures/empirical/tables/hyperparams_table}

\end{table*}

\begin{figure*}
    \centering
    \begin{subfigure}[b]{0.48\linewidth}
        \includegraphics[width=\linewidth]{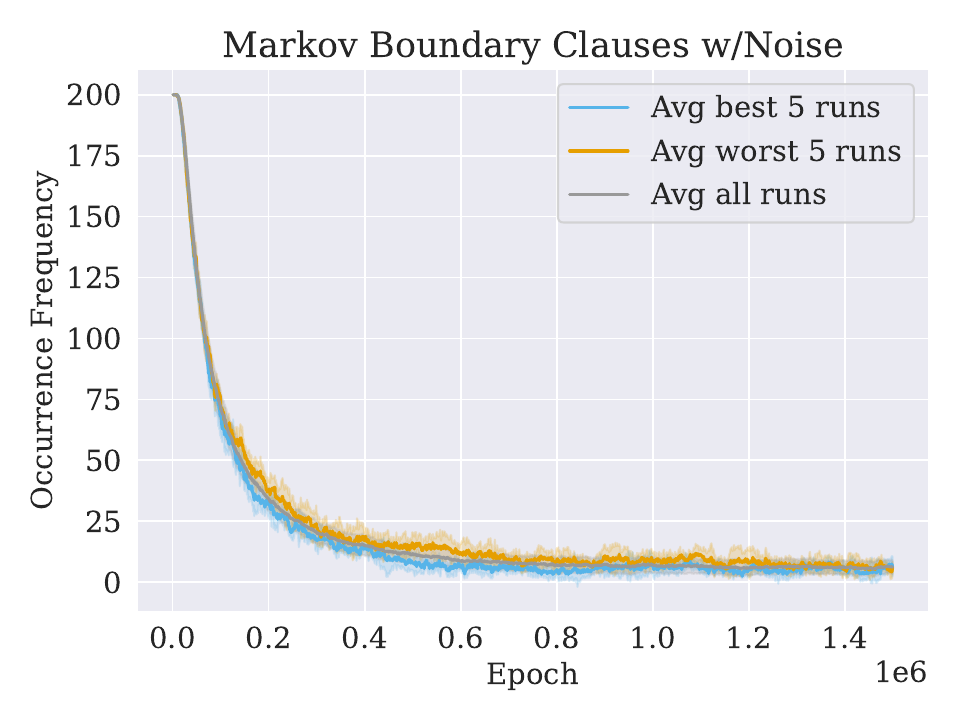}
        \caption{}
        \label{fig:markov-boundary-toy-1}
    \end{subfigure}
    \hfill
    \begin{subfigure}[b]{0.48\linewidth}
        \includegraphics[width=\linewidth]{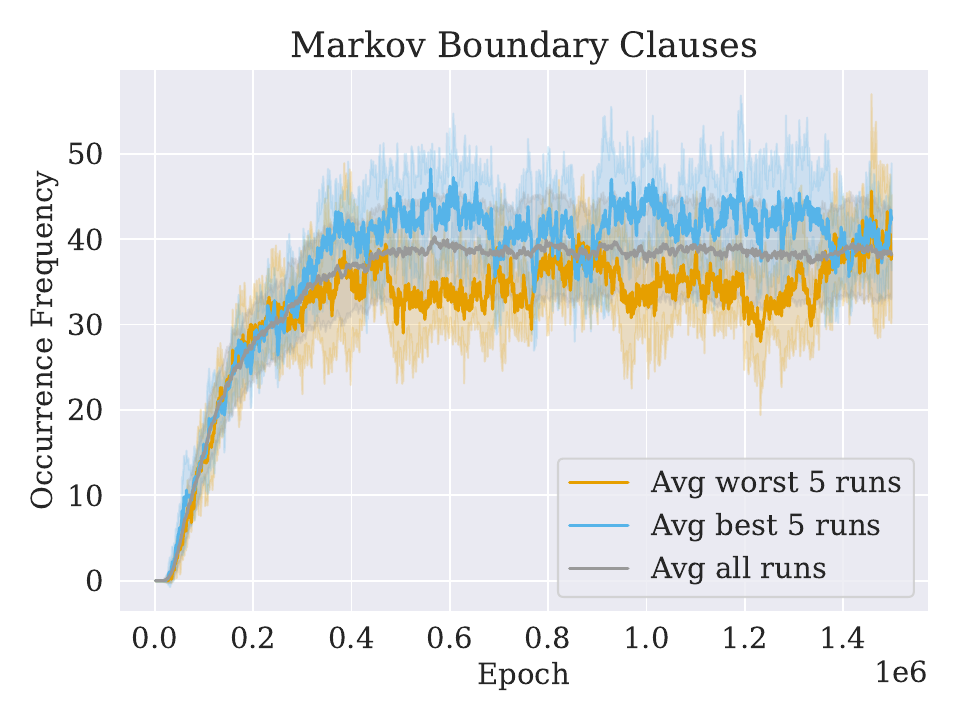}
        \caption{}
        \label{fig:markov-boundary-toy-2}
    \end{subfigure}
    \newline
    \begin{subfigure}[b]{0.48\linewidth}
        \includegraphics[width=\linewidth]{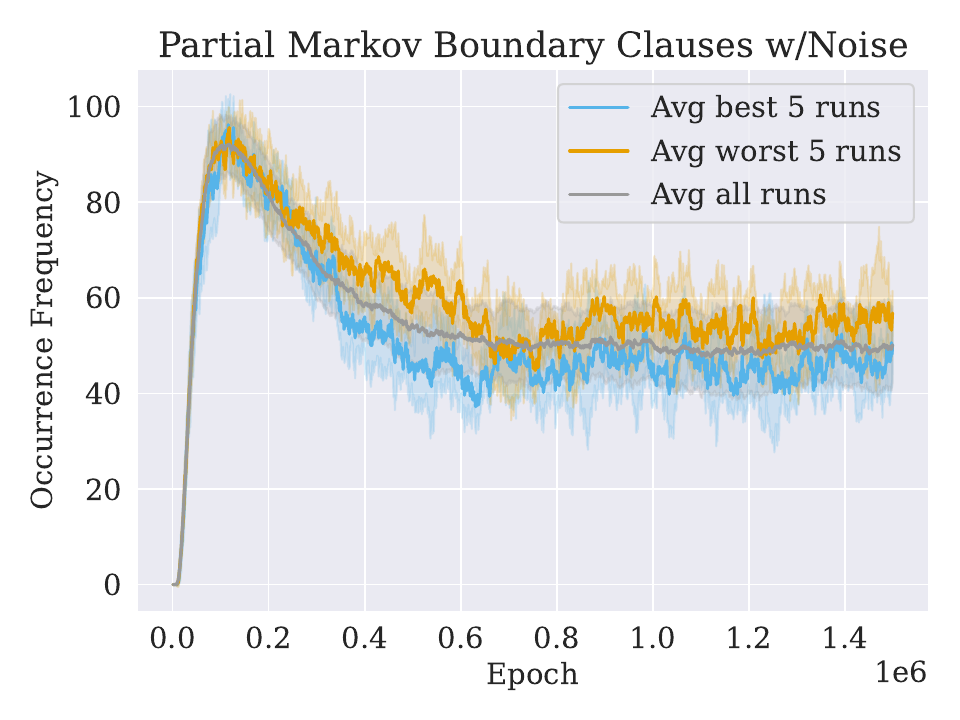}
        \caption{}
        \label{fig:markov-boundary-toy-3}
    \end{subfigure}
    \hfill
    \begin{subfigure}[b]{0.48\linewidth}
        \includegraphics[width=\linewidth]{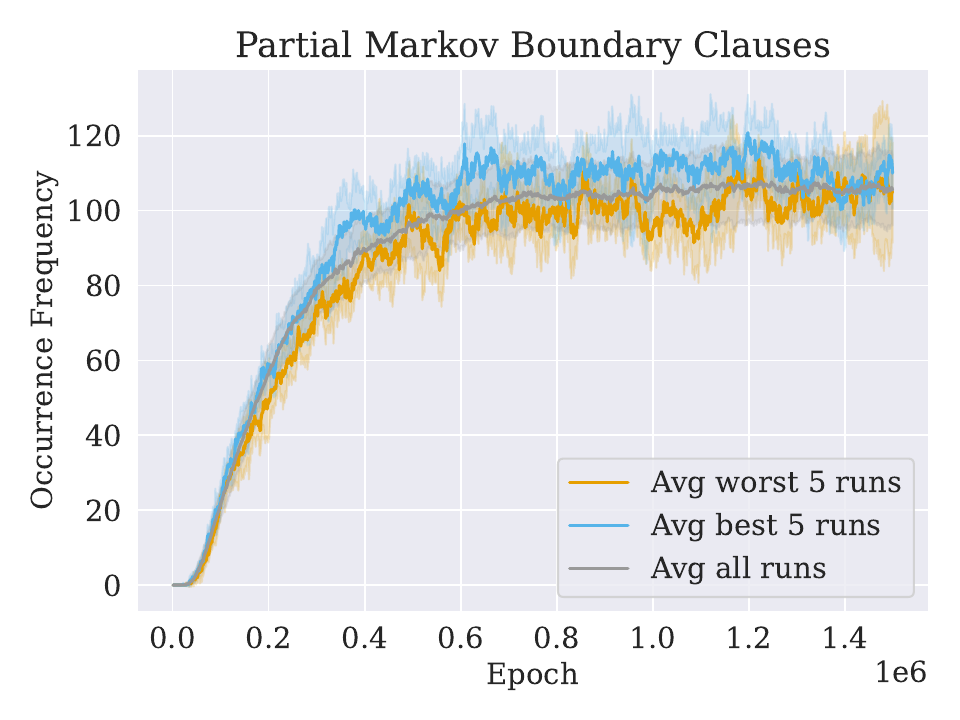}
        \caption{}
        \label{fig:markov-boundary-toy-4}
    \end{subfigure}
    \newline
    \caption{Distribution of Markov boundary among clauses. (a) Complete Markov boundary with supplementary noisy variables, (b) Complete Markov boundary without noisy variables, (c) Incomplete Markov boundary incorporating noisy variables, (d) Incomplete Markov boundary devoid of noise. The blue curve signifies the average of the top 5 performing trials, the orange curve represents the average of the bottom 5 trials, and the grey curve illustrates the mean value obtained from all 96 experimental runs.}
    \label{fig:markov-boundary-toy}
\end{figure*}

To examine Markov boundary-guided pruning empirically, we use the toy BN from Fig.~\ref{fig:BN} and Tables~\ref{tab:cpd-x5}-\ref{tab:cpd-y}. 

\subsection{Hyperparameters and experiment scope}
We conduct a hyperparameter search over \(T \in \{5, \ldots, 20\}\), \(s \in \{2, \ldots, 100\}\), and \(d \in \{20, \ldots, 400\}\). Furthermore, we explore number of Tsetlin automaton and CS-IA states,  \(N \in \{2^5, \ldots, 2^19\}\). Performance is assessed over 10 million epochs, incorporating 96 concurrent experiments sampled from the defined parameter space. For each epoch, the TM is trained using 100 samples from the toy BN illustrated in Fig. \ref{fig:BN}.\footnote{The code for conducting the experiments is publicly accessible at \url{https://github.com/cair/tmu}}

We report the best found hyperparameters in Table~\ref{tab:hyperparams}. Our main conclusion is that the CS-IA exhibits robustness towards varying hyperparameters, as demonstrated in Fig. \ref{fig:markov-boundary-toy} where best and worst performance is plotted across epochs over all the experiments.

\subsection{Literal occurrence with Markov boundary pruning}
We now investigate how accurately the Markov boundary is identified by examining how frequently the various literals appear within the TM clauses. As illustrated in Fig. \ref{fig:markov-boundary-literal-occurence}, Markov boundary-guided pruning effectively diminishes the presence of literals that do not exhibit causal relationships with the target variable \(Y\). In particular, \(X8\) distinguishes itself from the other literals by being left out of more clauses.

Also notice how the entropy of the CPDs (Table \ref{tab:cpd-x5}-\ref{tab:cpd-x4}) is reflected in the plot. Variables with high entropy (high uncertainty) are included in correspondingly fewer clauses. Variable \(X1\) meets two criteria: (1) it displays minimal entropy (Table \ref{tab:cpd-x1}), and (2) it constitutes an essential element of the Markov boundary. Consequently, this variable is expected to have a significant presence across all clauses of the TM. In contrast, variable \(X4\) is (1) included within the Markov boundary, but exhibits (2) substantial entropy. As a result, the presence of \(X4\) is considerably reduced following the training process. These findings indicate that the TM learning process produces a distribution of clauses that reflects the entropy of the underlying BN.

\subsection{Markov-boundary clause distribution}
We next explore how the clauses evolve during training, focusing on two aspects: (1) clauses containing the complete Markov boundary and (2) clauses with a partial Markov boundary, and their relationship to noisy variables. Our findings are illustrated in Fig. \ref{fig:markov-boundary-toy}.

Notice first in Fig. \ref{fig:markov-boundary-toy-3} that the clauses containing independent literals first increase, but then drops again after 100,000 epochs. This is due to the CS-IA getting sufficient confidence to start pruning literals. Eventually, 100 out of 200 clauses contain Markov boundary variables only, as shown in Fig. \ref{fig:markov-boundary-toy-4}.

The TM's capacity to form clauses with the complete Markov boundary is investigated in Fig. \ref{fig:markov-boundary-toy-2}, where approximately 40 out of 200 clauses are identified as complete Markov boundary clauses. Very few of these clauses contain non-boundary variables, as seen in Fig. \ref{fig:markov-boundary-toy-1}. Indeed, the Markov boundary-guided pruning achieves 140 out of 200 noise-free clauses, paving the way for a TM that learns causal relationships instead of correlations.

\subsection{Analysis of clause pruning for MNIST}

We finally study the effect of Markov boundary-based pruning on the more complex MNIST dataset by contrasting clause size against accuracy during learning. If Type III Feedback successfully prunes literals outside the Markov boundary, we expect reduced clause size without significantly affecting accuracy. Fig. \ref{fig:mnist} supports our hypothesis. The left figure shows behaviour without Type III Feedback, with average literals per clause growing to around $40$. In the right figure, Type III Feedback starts pruning literals around epoch 30, reducing the average number of literals per clause to below $5$, while maintaining accuracy. This indicates that superfluous parts of the clauses are being pruned.

\section{Theoretical analysis}
To analyze the proposed strategy, we study a toy-like example analytically. In more detail, we consider three variables, target variable $Y$, and two input variables $X_1$ and $X_2$. The Markov blanket of $Y$ is $X_1$, i.e., $P(Y | X_1, X_2) = P(Y | X_1)$. However, without knowing $X_1$, $X_2$ becomes informative: $P(Y | X_2) \neq P(Y)$. We assume that the clause of the TM $C$, after a period of training, becomes $C = X_1 \wedge X_2$ for predicting $Y=1$. We want to show that the Type~III Feedback can eventually guide the clause to become $C=X_1$.

\subsection{Probability of the steps being triggered}
Before we analyzing the convergence, we explain the meaning of ``Wait" in the learning loop as described in Sub-section~\ref{newLA}. In the learning loop, there are mainly two steps, namely, Step 1) and Step 2). Those two steps are controlled by the program and occurs in sequence, taking turns.  For example, the events of Step 1) are depicted as blacked dash lines in Fig.~\ref{figure:illustration}, index as 1, 2, 3, and 4. In the above example, when we study $X_1$, the event probability is described by $P(C=1, X_1=1)$.  Following the same concept, the events of Step 2) are presented by red solid lines, index as 5, 6, 7, 8. In the above example, when $X_1$ is studied, the event probability is described by $P(X_2=1)$. In the training process, without any control, those two types of events will not always take turn to happen, as shown in Fig.~\ref{figure:illustration}. When consecutive events from the same step occurs, the program only processes the first time when it happens and ignore latter consecutive ones, until an event from another step happens. In this concrete example shown in Fig.~\ref{figure:illustration}, the Event 1 is trigger for Step 1), but not Event 2. Similarly, Event 5 will be triggered, but not Event 6.  This applying to all previous and future events. Following the concept of taking turns, in Fig~\ref{figure:illustration}, Event 1, Event 3 and Event 4 will be triggered for Step~1). Similarly, Event 5 and Event 7 will be triggered for Step~2). 

Now let us study the probability of the events that will trigger Step 1) and Step 2) in turn, as described in Sub-section~\ref{newLA}. Clearly, Step 1) and Step 2) will be triggered with equal probability as they take turns to happen. However, to calculate the concrete probability in a closed form is difficult. Here, instead of calculating the event probability, we try to bound it. 

For $X_1$, let us define $P(\phi)$ the probability of Step 1) is triggered by the program. As Step 1) and Step 2) take turn to happen, the probability of Step 2) is also  $P(\phi)$. Clearly,  $P(\phi)\leq\min\{P(C=1, X_1=1),P(X_2=1)\}$. This is because when the two events take turn to be triggered, the event that has the minimum probability to be triggered determines the overall probability. In addition, for the case with the minimum probability, it is not all events that can trigger the program because the consecutive following ones are ignored. To simplify $\min\{P(C=1, X_1=1),P(X_2=1)\}$, we look at $P(C=1, X_1=1)=P(C=X_1\wedge X_2=1,X_1=1)=P(X_2=1,X_1=1)=P(X_2=1)P(X_1=1|X_2=1)\leq P(X_2=1)$. Therefore,  $P(\phi)\leq\min\{P(C=1, X_1=1),P(X_2=1)\}=P(C=1, X_1=1)=P(X_2=1,X_1=1)$.   

Similarly, for $X_2$, let us define $P(\phi')$ the probability of Step 1) is triggered by the program. As Step 1) and Step 2) take turn to happen, the probability of Step 2) is also  $P(\phi')$. Clearly,  $P(\phi')\leq\min\{P(C=1, X_2=1),P(X_1=1)\}$. Here $P(C=1, X_2=1)=P(C=X_1\wedge X_2=1,X_2=1)=P(X_2=1,X_1=1)=P(X_1=1)P(X_2=1|X_1=1)\leq P(X_1=1)$. Therefore,  $P(\phi')\leq\min\{P(C=1, X_2=1),P(X_1=1)\}=P(C=1, X_2=1)=P(X_2=1,X_1=1)$. Note that although $P(\phi')$ and $P(\phi)$ have the same bounds, the actual values may be quite different.

\subsection{The convergence of $X_1$}
We start the analysis of $X_1$ in different steps. For Step 1 (a), for $X_1$, the transition probability is $P(Y=1|C=1,X_1=1)=P(Y=1|C=X_1\wedge X_2=1,X_1=1)=P(Y=1|X_2=1,X_1=1)=P(Y=1|X_1=1)$, which is increasing. For step 1 (b), for $X_1$, $P(Y=0|C=1,X_1=1)=P(Y=0|C=X_1\wedge X_2=1,X_1=1)=P(Y=0|X_2=1,X_1=1)=P(Y=0|X_1=1)$, which is decreasing.  For step 1 (c), we have a decreasing probability, which is $d$. Those events are all conditioned by $P(\phi)$. 

We now study $X_1$ in Step 2.  For Step 2, when we study $X_1$, we need to ``Wait for the remaining variables of $C$ being 1". In this step, we need to wait until $X_2$ is 1. Then, we need to ignore the value of $X_1$. In this step, the value of $C$ is useless in the calculation. 
It only depends on the remaining variable, i.e., $X_2$ and the value $Y$. For Step 2 (a), namely when $Y=1$, the probability of decrease is $P(Y=1| X_2=1)$. For step 2 (b), namely, when $Y=0$, the probability of increase is $P(Y=0| X_2=1)$. Those events are all conditioned by $P(\phi)$ as well. 

Based on the above analysis, we can determine the probability of increase and the probability of decrease.
For increase, we have 1 (a) and 2 (b). Therefore, the overall probability of increase, defined by $P_{incX_1}$, is $P_{incX_1}=P(Y=1|X_1=1)P(\phi)$+$P(Y=0| X_2=1)P(\phi)$. For decrease, we have 1 (b), 1 (c) and 2 (a). Therefore, the overall decreasing probability, defined by $P_{decX_1}$ becomes $P_{decX_1}=P(Y=0|X_1=1)P(\phi)+dP(\phi)+P(Y=1| X_2=1)P(\phi)$. Fig.~\ref{figure:TAarchitecture_basic} illustrates the transitions of the states. Here $P_{incX_1}$ is represented by the dashed arrow while $P_{decX_1}$ is presented by the solid arrow. Note here that $P_{decX_1}\leq1$ and $P_{incX_1}\leq1$ must fulfill and the self-loop is not depicted. This condition is not difficult to fulfill as long as the dateset is balanced and the value $d$ is very small.  

Now let us study the infinite time horizon with infinite states in the TA. As shown by Lemma 1 in~\cite{zhang2020convergence}, $X_1$ is to be included if  $P_{decX_1}<P_{incX_1}$ holds in probability 1. To full fill this requirement, we need
\begin{align}
&P(Y=1|X_1=1)+P(Y=0|X_2=1)\nonumber\\
&>P(Y=0|X_1=1)+d+P(Y=1| X_2=1), 
\end{align}
which can be re-written as:
\begin{align}
&P(Y=1|X_1=1)-P(Y=0|X_1=1)\nonumber\\
&-\left(P(Y=1| X_2=1)-P(Y=0|X_2=1)\right) >d. 
\label{Req_X_1}
\end{align}
Eq.~(\ref{Req_X_1}) can be interpreted in this way. Given $X_1=1$, the probability difference between $Y$ being 1 and 0 is $P(Y=1|X_1=1)-P(Y=0|X_1=1)$. Similarly, for $X_2=1$, the probability difference between $Y$ being 1 and 0 is $P(Y=1|X_2=1)-P(Y=0|X_2=1)$. When $X_1$ is not pruned, meaning that $X_1$ can be used to predict $Y$ by itself (without information from $X_2$), i.e., $C=X_1$, we expect that given $X_1=1$, the probability of $Y$ being 1 is greater than that of being 0. In addition, the difference given $X_1$ must be greater than that given $X_2$, meaning that the information of $X_2$ is redundant.  A margin $d$ is given to ensure that the above mentioned differences are sufficient. 

\subsection{The convergence of $X_2$} In this subsection, we analyze the behavior of $X_2$. For Step 1) (a), the transition probability is $P(Y=1|C=1,X_2=1)=P(Y=1|C=X_1\wedge X_2=1,X_2=1)=P(Y=1|X_2=1,X_1=1)=P(Y=1|X_1=1)$, which is increasing. For step 1 (b), for $X_2$, $P(Y=0|C=1,X_1=1)=P(Y=0|C=X_1\wedge X_2=1,X_2=1)=P(Y=0|X_2=1,X_1=1)=P(Y=0|X_1=1)$, which is decreasing.  For step 1 (c), we have a decreasing probability, which is $d$. Those events are all conditioned by $P(\phi')$.

We now study $X_2$ in Step 2.  For Step 2, when we study $X_2$, we need to ``Wait for the remaining variables of $C$ being 1". In this step, we need to wait until $X_1$ is 1. Then, we need to ignore the value of $X_2$. In this step, the value of $C$ is useless in the calculation. 
It only depends on the remaining variable, i.e., $X_1$ and the value $Y$. For Step 2 (a), namely when $Y=1$, the probability of decrease is $P(Y=1| X_1=1)$. For step 2 (b), namely, when $Y=0$, the probability of increase is $P(Y=0| X_1=1)$. Those events are all conditioned by $P(\phi')$ as well.

Based on the above analysis, we can determine the probability of increase and the probability of decrease.
For increase, we have 1 (a) and 2 (b). Therefore, the overall probability of increase, defined by $P_{incX_2}$, is $P_{incX_2}=P(Y=1|X_1=1)P(\phi')$+$P(Y=0| X_1=1)P(\phi')$. For decrease, we have 1 (b), 1 (c) and 2 (a). Therefore, the overall decreasing probability, defined by $P_{decX_2}$ becomes $P_{decX_2}=P(Y=0|X_1=1)P(\phi')+dP(\phi')+P(Y=1| X_1=1)P(\phi')$. Fig.~\ref{figure:TAarchitecture_basic} illustrates the transitions of the states. Here $P_{incX_2}$ is represented by the dashed arrow while $P_{decX_2}$ is presented by the solid arrow. Note here that $P_{decX_2}\leq1$ and $P_{incX_2}\leq1$ must fulfill and the self-loop is not depicted. This condition is not difficult to fulfill as long as the dateset is balanced and the value $d$ is very small. 

In infinite time horizon, given infinite length of Markov chain, to make sure that $X_2$ is to be pruned, we have to grantee that $P_{decX_2}>P_{incX_2}$. This is always true as long as $d>0$ holds. 

To conclude, based on the above analysis, we can see that for this toy-like example, if the system configuration is correct, i.e.,  $P(Y=1|X_1=1)-P(Y=0|X_1=1)-\left(P(Y=1| X_2=1)-P(Y=0|X_2=1)\right)>d>0$, the system will almost surely converge to the correct clause with proper pruning and keeping, i.e., arriving $C=X_1$, given infinite time and states. 

\section{Conclusion and Further Work}

In this paper, we proposed a novel TM feedback scheme that supplements Type I and Type II Feedback. The new Type III Feedback exploits context-specific independence analysis, manifested as a Context-Specific Independence Automaton. The automaton evaluates independence hypotheses on-line, allowing it to uncover literals that are independent of the prediction target, given the remaining literals of the clause (the context).

Our empirical and theoretical analysis shows that the scheme is effective, opening up for further synergy between the fields of BNs and TMs. In our future work, we intend to investigate how TMs can enable learning of Bayesian knowledge bases~\cite{Santos2011TuningAB} and efficient probabilistic inference by composing Markov-boundary pruned TM clauses into, e.g., a Moral Graph~\cite{cowell1999}.

\bibliographystyle{IEEEtran}  
\bibliography{References}

\end{document}

%% file: Figures/empirical/tables/hyperparams_table.tex
\begin{tabular}{rrrrrrrr}
\toprule
T & s & d & State bits TA & State bits IA & Weighted & MB Clauses & OR \\
\midrule
9.45 & 45.74 & 83.36 & 14 & 20 & False & 128.26 & 40000 \\
6.22 & 23.74 & 336.11 & 10 & 14 & True & 119.49 & 47000 \\
12.41 & 78.24 & 390.94 & 14 & 14 & True & 119.59 & 37000 \\
\bfseries 17.33 & \bfseries 66.81 & \bfseries 226.18 & \bfseries 6 & \bfseries 10 & \bfseries True & \bfseries 129.12 & \bfseries 35000 \\
12.72 & 53.98 & 146.52 & 12 & 10 & False & 119.52 & 55000 \\
16.37 & 67.13 & 119.71 & 20 & 16 & True & 127.57 & 46000 \\
\bottomrule
\end{tabular}
\caption{Hyperparameters corresponding to the most effective and least effective runs of the toy experiment. The evaluation criteria is the number of clauses that have successfully pruned variables situated outside the Markov boundary. The best discovered configuration is in bold. The OR columns signify the timestep at which clauses containing a partial Markov Boundary jointly find the complete boundary by ORing the partial ones.}
\label{tab:hyperparams}